\documentclass{article}
\usepackage{ijcai26}

\usepackage{times}
\usepackage{soul}
\usepackage{url}
\usepackage[hidelinks]{hyperref}
\usepackage[utf8]{inputenc}
\usepackage[small]{caption}
\usepackage{capt-of}
\usepackage{graphicx}
\usepackage{amsmath}
\usepackage{amssymb}
\usepackage{amsthm}
\usepackage{booktabs}
\usepackage{float}
\usepackage[switch]{lineno}

\title{Forgetting is Not Erasure: Recovering Latent Knowledge via Transport Keys}

\author{
Archie Chaudhury$^1$
\affiliations
$^1$Axionic Labs\\
\emails
archie@axioniclabs.ai
}

\begin{document}

\maketitle

\begin{abstract}
Catastrophic forgetting is often framed as a representational problem: after sequential training, a model appears to lose the features that supported performance on earlier tasks. We challenge the stronger form of this view. Across controlled continual-learning settings, we find that a significant portion of apparent forgetting can be attributed to interface drift between internal stages rather than permanent erasure of task-relevant computation. We study this phenomenon through a stitched evaluation protocol that combines early computation from a post-update network with late computation from its predecessor, optionally mediated by a compact, task-specific transport key. We describe transport keys at a systems level as compact interface-alignment operators estimated from a small set of paired anchor activations and evaluated through model stitching. On split CIFAR-100 with a ResNet-style network, transport keys recover most of the original Task A performance after sequential training on Task B. On a compact vision transformer, we observe a similar recovery pattern. These results suggest that continual learning may require better mechanisms for indexing and re-accessing latent computations, not only methods that prevent weight change.
\end{abstract}

\section{Introduction}
Continual Learning (CL) refers to the ability of an autonomous agent or machine to learn multiple different types of knowledge and behave as if it learned all of them at once \cite{wang2024comprehensivesurveycontinuallearning}. Enabling CL remains one of the largest problems in contemporary deep networks, where fine-tuning or training on a new task often results in the collapse of accuracy or performance on a prior task. This collapse has been historically defined as catastrophic forgetting, or erasure, where a model's ability to generalize is limited by its internal representations \cite{MCCLOSKEY1989109}. Catastrophic forgetting is often framed as the result of a fundamental, architectural compromise between stability and plasticity: the only way to enable a model to learn new tasks without losing performance on prior tasks is to scale it \cite{FRENCH1999128}. 
In this work, we show that in high-dimensional neural representations, forgetting can be modeled as an \textbf{access problem}, being the result of drift between various network stages, and that training on a new task need not necessarily result in the permanent erasure of a model's ability to do a task it observed previously. By modeling a neural network's representation of a new learned task as a staged computation where later layers learn to interpret intermediate activations, we define forgetting as a dimensional problem, in which the internal features of a model are altered significantly when learning a new task, leaving the original set of features associated with a task invalid. Under this view, a model's performance on a task or functionality over a wide range of scenarios can be recovered through the use of a small key that realigns its internal features at specific internal interfaces. Drawing upon model stitching \cite{Bansal2021Stitching}, we construct such a key, allowing us to recover latent task-specific features and lost performance.
\paragraph{Contributions}
Our specific contributions are as follows.

(i) We present evidence that a substantial component of catastrophic forgetting can be interpreted as an interface-access failure rather than complete representational erasure.

(ii) We introduce the transport-key framing: a compact, task-specific alignment object that restores compatibility between stages of sequentially trained networks by correcting activation-space interface drift.

(iii) We provide a stitched evaluation protocol that directly tests whether post-update networks preserve task-relevant latent computation, and we use controls to distinguish genuine interface alignment from generic adaptation.

(iv) We report preliminary results on ResNet-style and transformer-style vision models showing substantial recovery of Task A performance after sequential training.

\paragraph{Related Work}
Continual learning (CL) has been studied extensively across supervised and reinforcement learning settings, with a focus on addressing the inherent tradeoff between long-term stability and plasticity \cite{FRENCH1999128}. Contemporary approaches to addressing forgetting have mainly been focused on constraining weight updates to specific parameters, such as Elastic Weight Consolidation (EWC) \cite{Kirkpatrick2017EWC}, Synaptic Intelligence (SI) \cite{Zenke2017SI}, and Memory Aware Synapses \cite{Aljundi2018MAS}. Alternative methodologies have leaned toward dynamically generating  prior task examples while training on new ones, as seen in iCaRL \cite{Rebuffi2017iCaRL} and Experience Replay \cite{Rolnick2019ER}. Meanwhile, gradient-based methods such as GEM \cite{LopezPaz2017GEM} and A-GEM \cite{Chaudhry2019AGEM} enforce guards that ensure specific weight updates do not affect the features associated with prior tasks.

We implement the stitching evaluation as an extension of model stitching, initially defined by \cite{LencVedaldi2015,Bansal2021Stitching}. Traditionally, stitching has been utilized to compare architectures or to merge models in weight-space \cite{Entezari2022,Ainsworth2023}. We adapt stitching to a continual learning setting by treating the network post-training as the sender and its predecessor as the receiver. As such, we are able to test directly for the existence of latent knowledge, without needing retraining or gradient updates, in contrast to the parameter-based methods such as those implemented by \cite{Rusu2016ProgNN,Mallya2018PackNet}, or adapter-dependent methods such as \cite{Houlsby2019Adapters}.

\section{Background and Formalization}
We start by defining some preliminaries, particularly centered around formalizing sequential training as it applies to a standard CL setting. We also provide a generalized definition of representational drift, which is essentially a value that quantifies the difference in a model's ability to perform a Task\textbf{A} once it has been subsequently trained on a different Task\textbf{B}.

\subsection{Sequential Training}

We start with a standard neural network $f(x; \theta)$ that maps inputs $x$ to outputs, and train it on a sequential set of tasks.  For simplicity, we restrict our focus to two tasks, although in practice, this extends to any number of tasks, all trained one after the other.

Let $\theta_A$ denote the parameters after training on an individual starting Task A. We then continue training on Task B, yielding parameters $\theta_{AB}$. In classic CL, this results in the accuracy on task A for the model with parameters $\theta_{AB}$ degrading significantly, with the most common explanation being that the process of training on Task B overwrites the weights that defined the features associated with Task A.

\subsection{Network Decomposition}
Contemporary deep networks are essentially a broad combination of different stages. For example, in a traditional ResNet-style network, these stages are essentially the core stem, and then four residual blocks that get progressively deeper. 

At any individual stage $\ell$, we can split the network into two parts:
\begin{itemize}
    \item The early network $f_{\leq \ell}$, which maps an input $x$ to an intermediate activation tensor $h_\ell \in \mathbb{R}^{C \times H \times W}$.
    \item The late network $f_{>\ell}$, which maps $h_\ell$ to a prediction, including the classification head.
\end{itemize}

This decomposition allows us to precisely determine the stage at which performance on Task A drops once the network's parameters have been updated.

\subsection{Interface Drift}

Our primary hypothesis is that the catastrophic forgetting often seen in neural networks occurs not at a particular stage, but at the interface, such that the later stages in a network are no longer able to recognize the features for a prior task produced by the earlier stages.

To formalize this, suppose there exists a transformation $T_\ell$ such that
\begin{equation}
    h_\ell(x; \theta_A) \approx T_\ell\bigl(h_\ell(x; \theta_{AB})\bigr)
\end{equation}
for inputs $x$ from Task A. If $T_\ell$ is simple, then the network post-update is still able to compute and encode information associated with task A: the problem arises during decoding, when the later stages of the network (which have now been updated) expect a different computational result for the features it is decoding.

\begin{figure}[H]
    \centering
    \includegraphics[width=\linewidth]{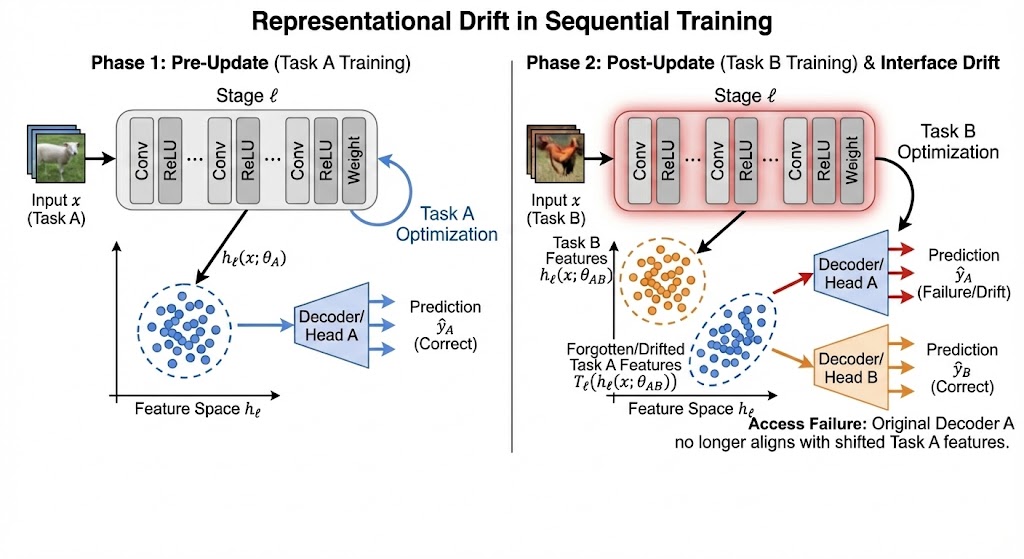}
    \caption{Comparison of traditional erasure versus interface drift}
    \label{fig:interface_drift}
\end{figure}

Figure~\ref{fig:interface_drift} showcases the phenomenon.
Rather than the features associated with Task A being permanently deleted as a result of training on Task B, they are simply uninterpretable by the later stages of the network. We call this phenomenon \textbf{interface drift}. 

\section{Methodology}
\label{sec:methodology}

This section explains transport keys at the level needed to understand the mechanism and the empirical claim. The central object is an interface-level alignment operator: it is inserted between an early portion of a post-update network and a late portion of the pre-update network, and it maps the new activation coordinate system back into a form that the old downstream computation can decode.

\subsection{Overview: Transport Keys as Interface Alignment}

Let $f(x;\theta)$ be a neural network trained first on Task A and then on Task B. Let $\theta_A$ denote the parameters after Task A training and $\theta_{AB}$ denote the parameters after subsequent Task B training. At an internal stage $\ell$, we decompose the network into an early map $f_{\leq \ell}$ and a late map $f_{>\ell}$. For an input $x$ from Task A, the two checkpoints induce paired activations
\begin{equation}
    h^A_\ell(x) = f^A_{\leq \ell}(x), \qquad
    h^{AB}_\ell(x) = f^{AB}_{\leq \ell}(x).
\end{equation}
A transport key $T_\ell$ is a compact transformation that acts on $h^{AB}_\ell(x)$ so that the transformed activation is readable by the old Task A decoder:
\begin{equation}
    \widetilde{h}^{A}_\ell(x) = T_\ell(h^{AB}_\ell(x)).
\end{equation}
Rather than being a new trained object or replay buffer, a transport key is simply an activation-space alignment object attached to a particular interface. Its role is to correct the mismatch between what the post-update early network now emits and what the pre-update late network expects to receive.

Intuitively, sequential training can change an internal representation in at least two ways. First, it can shift or rescale individual channels, leaving the semantic content mostly intact but changing calibration. Second, it can rotate or mix features across channels, leaving the information present but expressed in a different basis. Transport keys are designed to correct these interface-level changes while preserving the original downstream decision rule.

\subsection{Anchor Sets}

To estimate a key, we use a small anchor set drawn from Task A. Anchors are ordinary examples from the earlier task and are passed through both checkpoints. This produces paired activations at the same interface: one activation from the Task A checkpoint and one from the post-update checkpoint. The use of paired anchors is important because the key is intended to align corresponding computations, not merely match aggregate activation statistics.

Balanced anchor selection is used when the earlier task is a classification problem, so that the alignment signal is not dominated by a small subset of classes. The anchor set is used only to build the interface key. It does not retrain the backbone, does not update the Task A head, and does not introduce new labels at evaluation time.

\subsection{Key Families}

We consider two conceptual families of transport keys. The first is a channel-calibration key. It corrects per-channel drift, such as changes in activation scale or offset. This form is deliberately small and is most effective when sequential training preserves the original channel basis but changes its calibration. In the same-domain CIFAR experiments, this compact correction explains most of the recovered accuracy.

The second is a cross-channel alignment key. It allows structured mixing between channels and is useful when the post-update representation has changed basis more substantially. This form is still applied at the activation interface rather than through end-to-end retraining. It becomes more important under domain shift, where a network trained after CIFAR-10 on SVHN may preserve useful CIFAR-10 information but express it in a less directly compatible coordinate system.

These key families correspond to two interpretable modes of interface drift: \emph{calibration drift}, where individual channels remain meaningful but change scale, and \emph{mixing drift}, where information is distributed across channels in a different basis. The experiments below use this distinction to explain why small keys suffice in some settings while more expressive keys matter in others.

\subsection{Stitched Evaluation}

We evaluate transport keys using model stitching. Given the post-update early network $f_{\leq \ell}^{AB}$, the pre-update late network $f_{>\ell}^{A}$, and a transport key $T_\ell$, the stitched model is
\begin{equation}
    (f_{\leq \ell}^{AB} \oplus_{T_\ell} f_{>\ell}^{A})(x)
    = f_{>\ell}^{A}\bigl(T_\ell(f_{\leq \ell}^{AB}(x))\bigr).
\end{equation}
For Task A evaluation, the stitched prediction is
\begin{equation}
    \hat{y}=\mathrm{head}_A\bigl(f_{>\ell}^{A}(T_\ell(f_{\leq \ell}^{AB}(x)))\bigr).
\end{equation}
The Task A head is preserved and is not retrained. This design makes the evaluation strict: recovery must come from restoring compatibility with the old internal decoder, not from learning a fresh classifier on top of post-update features.

This protocol separates three cases. If the post-update early network has destroyed the information needed for Task A, keyed stitching should fail. If the information remains but has drifted out of the coordinate system expected by the old decoder, no-key stitching should perform poorly while keyed stitching should recover. If the interface remains mostly compatible, no-key stitching itself can recover substantial performance.

\subsection{Controls}

We use controls to test whether the key is exploiting real interface structure. The no-key control sends $h^{AB}_\ell$ directly into the old downstream network. Channel-disruption controls test whether channel identity matters for compact calibration keys. Correspondence-breaking controls test whether the mapping depends on paired examples rather than only marginal statistics. Stage-wise controls evaluate the same procedure at multiple internal boundaries.

The expected signature is specific: valid keyed stitching should outperform the forgotten model and the no-key stitch; disrupting channel structure or example correspondence should reduce recovery; and recovery should vary systematically by stage. This is the pattern observed in the experiments below.

\begin{figure}[t]
    \centering
    \includegraphics[width=\linewidth]{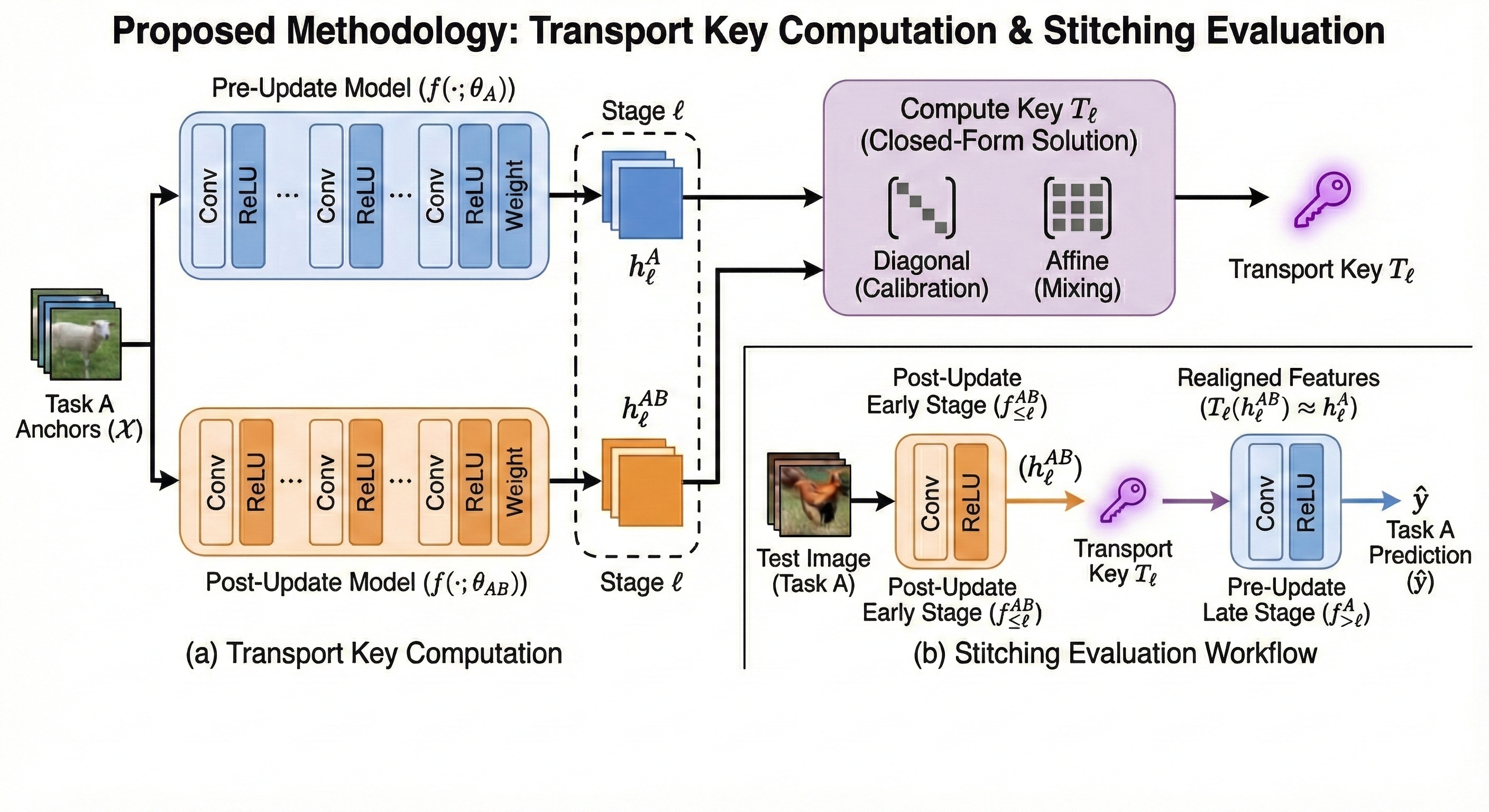}
    \caption{Transport-key evaluation workflow. Paired anchor activations from the Task A checkpoint and the post-update checkpoint are used to build an interface key. The key is then inserted into a stitched model that combines post-update early computation with the original Task A downstream decoder.}
    \label{fig:methodology_workflow}
\end{figure}

\section{Experimental Setup}

We evaluate the transport-key hypothesis in controlled image-classification continual-learning settings. The purpose of these experiments is not to establish a production-ready continual-learning system, but to test whether post-update networks retain latent computations that can be re-accessed by an internal alignment mechanism.

\subsection{Datasets and Task Constructions}

We use CIFAR-100, CIFAR-10, and SVHN as standard vision benchmarks. CIFAR-100 enables same-domain task splits; CIFAR-10 to SVHN provides a stronger domain-shift test because natural-object recognition is followed by digit recognition. We consider four settings, summarized in Table~\ref{tab:experiments}: a two-task CIFAR-100 split, a longer three-task CIFAR-100 sequence, a CIFAR-10 to SVHN domain-shift sequence, and a compact vision-transformer variant of the CIFAR-100 split.

\begin{table}[t]
\centering
\resizebox{\columnwidth}{!}{%
\begin{tabular}{@{}lcccc@{}}
\toprule
Setting & Tasks & Classes & Domain & Architecture \\
\midrule
Split CIFAR-100 & A $\to$ B & 50 / 50 & Same & ResNet-style \\
Three-Task & A $\to$ B $\to$ C & 33/33/34 & Same & ResNet-style \\
Domain Shift & C10 $\to$ SVHN & 10 / 10 & Diff. & ResNet-style \\
Transformer & A $\to$ B & 50 / 50 & Same & Mini ViT \\
\bottomrule
\end{tabular}%
}
\caption{Summary of experimental settings.}
\label{tab:experiments}
\end{table}

\subsection{Models and Training}

The main convolutional experiments use a ResNet-style network adapted to small images. The transformer experiment uses a compact ViT-style architecture. In each setting, the model is trained sequentially: first on Task A, then on one or more subsequent tasks. The Task A head is preserved for evaluation so that degradation in Task A performance is attributable to changes in the shared representation rather than replacement of the classifier.

The baseline sequential-training setup follows ordinary supervised continual-learning practice. Each model is trained on Task A, then continued on the subsequent task or tasks. Transport keys are computed after the sequential update from paired internal activations and are evaluated only through the stitched protocol described above. This keeps the recovery test separate from ordinary retraining or adapter fine-tuning.

\subsection{Evaluation Metrics}

Our primary metric is classification accuracy on the Task A test set. We report four quantities:

\begin{enumerate}
    \item \textit{Pre-update accuracy}: Task A accuracy immediately after Task A training, using parameters $\theta_A$.
    \item \textit{Post-update accuracy}: Task A accuracy after subsequent training, using the post-update shared network with the preserved Task A head.
    \item \textit{Stitched accuracy}: Task A accuracy using the stitched network defined in Section~\ref{sec:methodology}.
    \item \textit{Recovery rate}: the fraction of lost accuracy recovered by stitching,
    \begin{equation}
        \text{Recovery} = \frac{\text{Stitched} - \text{Post-update}}{\text{Pre-update} - \text{Post-update}}.
    \end{equation}
\end{enumerate}

A recovery rate of 100\% means the stitched network matches the original Task A accuracy. A recovery rate near 0\% means stitching provides no improvement over direct post-update evaluation.

\subsection{Stage Selection}

We evaluate stitching at multiple internal stage boundaries. Earlier stages test whether low-level and mid-level features remain accessible after sequential training. Later stages test whether more task-specific representations can be realigned. We report stage identifiers such as $s1$--$s4$ for convolutional networks and $b0$--$b3$ for transformer blocks.

\section{Results}

We present our results across all aforementioned experimental settings. Table~\ref{tab:main_results} summarizes the primary findings.

\begin{table}[t]
\centering
\small
\begin{tabular}{@{}llrrr@{}}
\toprule
Setting & Stage & Pre-update & Post-update & Keyed \\
\midrule
Split CIFAR-100 & $s1$ & 0.750 & 0.392 & 0.721 \\
Three-Task (A$\to$B$\to$C) & $s1$ & 0.735 & 0.337 & 0.674 \\
CIFAR-10$\to$SVHN & $s1$ & 0.877 & 0.154 & 0.752 \\
Mini ViT & $b0$ & 0.590 & 0.315 & 0.542 \\
\bottomrule
\end{tabular}
\caption{Task A accuracy before training on subsequent tasks (Pre-update), after training (Post-update), and after applying a transport key at the indicated stage (Keyed). All settings show substantial recovery.}
\label{tab:main_results}
\end{table}

\subsection{Split CIFAR-100}

\begin{figure}[t]
\centering
\includegraphics[width=0.95\linewidth]{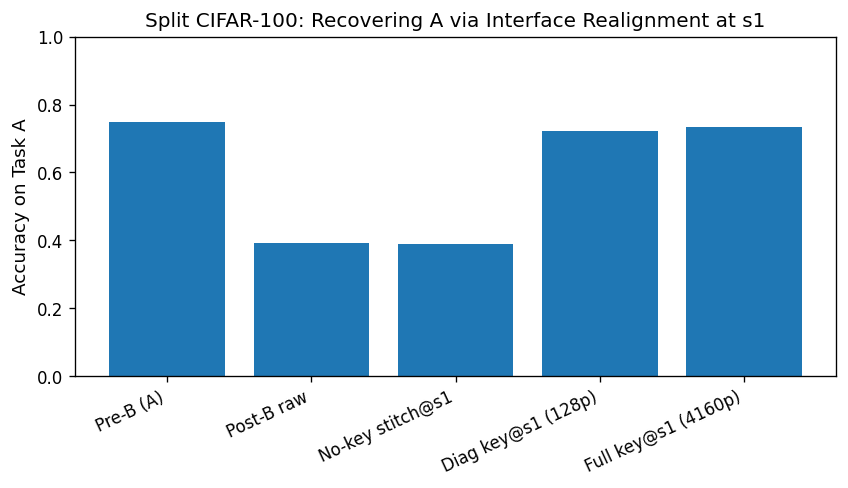}
\caption{Split CIFAR-100 recovery via stitching at stage 1. A compact transport key recovers 92\% of lost Task A accuracy. A more expressive cross-channel variant provides only marginal additional benefit.}
\label{fig:splitcifar100_main}
\end{figure}

Figure~\ref{fig:splitcifar100_main} showcases our primary result. After training on Task B, Task A accuracy drops from 0.750 to 0.392. No-key stitching at stage 1 yields 0.390, indicating that the interface is not already compatible. A compact transport key recovers accuracy to 0.721, restoring 92\% of lost performance. A more expressive cross-channel key family provides only marginal additional benefit, suggesting that the relevant drift in this setting is mostly structured channel-calibration drift rather than arbitrary feature rewriting.

The recovery degrades significantly at later stages. Table~\ref{tab:stage_sweep} shows that keyed stitching at $s1$ achieves 92\% recovery, dropping to 71\% at $s2$, 50\% at $s3$, and failing entirely at $s4$. This indicates that early features associated with a particular task do remain largely intact and encapsulate the majority of latent features.

\begin{table}[t]
\centering
\small
\begin{tabular}{@{}lccr@{}}
\toprule
Stage & No Key & Keyed & Recovery \\
\midrule
$s1$ & 0.390 & 0.721 & 92.0\% \\
$s2$ & 0.374 & 0.647 & 71.3\% \\
$s3$ & 0.285 & 0.464 & 50.3\% \\
$s4$ & 0.392 & 0.386 & $-$1.7\% \\
\bottomrule
\end{tabular}
\caption{Stage-wise recovery in Split CIFAR-100.}
\label{tab:stage_sweep}
\end{table}

\subsection{Extended Task Sequences}

The three-task experiment (A$\to$B$\to$C) tests whether interface drift is only concentrated on limited sequences. Task A accuracy degrades from 0.735 after initial training to 0.450 after Task B and 0.337 after Task C. Despite two updates, a transport key at $s1$ recovers accuracy to 0.674, restoring 85\% of lost performance. The stage-wise gradient replicates: $s1$ achieves 85\% recovery, falling to 71\% at $s2$, 50\% at $s3$, and failing at $s4$.

\subsection{Domain Shift}

In the domain-shift setting (CIFAR-10$\to$SVHN), the drop in performance is severe, with Task A accuracy collapsing from 0.877 to 0.154. Yet, no-key stitching at $s1$ already recovers to 0.619, indicating substantial latent retention even when the subsequent task involves a different visual domain. 

Table~\ref{tab:domain_controls} presents the control analysis. The transport key substantially improves accuracy at the earliest stage, while structure-disrupting controls reduce the gain. At the next stage, the same pattern becomes sharper: keyed recovery remains strong, while a correspondence-breaking control collapses close to the post-update baseline. These results support the claim that recovery depends on structured alignment rather than generic post-hoc adaptation.

\begin{table}[t]
\centering
\small
\begin{tabular}{@{}lrrrr@{}}
\toprule
Stage & Keyed & Control A & Keyed+ & Control B \\
\midrule
$s1$ & 0.705 & 0.332 & 0.752 & 0.536 \\
$s2$ & 0.501 & 0.400 & 0.644 & 0.184 \\
\bottomrule
\end{tabular}
\caption{Control analysis under domain shift (CIFAR-10$\to$SVHN). Perm is a channel-permutation control for compact calibration keys. Rand is a correspondence-breaking control for cross-channel keys.}
\label{tab:domain_controls}
\end{table}

\subsection{Architectural Generality}

The Mini ViT experiment tests whether interface drift is specific to convolutional architectures. Task A accuracy drops from 0.590 to 0.315 after Task B training. Stitching at block 0 with a transport key recovers to 0.542 (83\% recovery). Notably, no-key stitching already achieves 0.541, suggesting that early transformer representations are highly stable under fine-tuning, with drift concentrated in late blocks. The stage gradient replicates: recovery degrades from 83\% at $b0$ to 68\% at $b1$, 42\% at $b2$, and fails at $b3$.

\section{Discussion}

Our results show that forgetting in continual-learning settings is not necessarily permanent, and that substantial recovery can be achieved through a compact transport key. Across all settings, our stitched evaluation protocol showed that task-specific computations persist as latent features that are no longer directly accessible, rather than being permanently erased. This was most evident in the cross-domain experiment, where stitched evaluation recovered a large portion of the lost accuracy.

We also showed that forgetting is localized, and is primarily the result of representational drift between interfaces rather than erasure. In all experiments, transport keys applied at the early stages of the network were able to recover most of the pre-update performance, while keys at later stages were less effective. This suggests that early layers are more general, and often store latent features associated with individual, prior tasks.

Third, interface drift appears to have internal structure. The same-domain setting can be corrected with a highly compact key, while the domain-shift setting benefits from a more expressive cross-channel variant. The control experiments indicate that recovery depends on the relationship between corresponding examples and internal interfaces, not merely on aggregate activation statistics.

These findings show that forgetting is potentially an access problem, rather than being a representational one. Future systems in CL settings could store individual keys for each task, rather than needing to freeze their weights at a particular checkpoint or maintaining replay buffers. In our main experiment, a compact key was able to recover most lost accuracy in a ResNet-style network while remaining small relative to the base model. This suggests that there need not necessarily be a tradeoff between stability and plasticity; indeed, deep networks regardless of their size may be able to achieve a high degree of plasticity if they are able to maintain an index that allows them access knowledge of prior tasks.

\paragraph{Limitations and Future Work} We note several limitations in our work here. Our experiments only use a maximum of three tasks in an individual task for simplicity. Longer sequences may exhibit drift that is unable to be fixed by our lightweight transport keys. The stitched evaluation requires access to pre-update components, which makes it primarily a diagnostic rather than a deployment recipe. Future work will investigate in-place recovery and online variants that preserve the same access-based principle without requiring a full predecessor network at inference time.
\section*{Ethical Statement}
A relevant ethical consideration is that improved recovery of prior capabilities could be used to preserve or restore capabilities that a deployer intended to remove. We therefore treat transport-key systems as controlled research infrastructure. Broader release should be paired with safety evaluations, access controls, and clear restrictions on applications that restore harmful or intentionally deprecated capabilities.

\bibliographystyle{named}
\bibliography{ijcai26}

\appendix

\section{Anchor Efficiency on Split CIFAR-100}

\begin{figure}[h]
    \centering
    \includegraphics[width=0.85\linewidth]{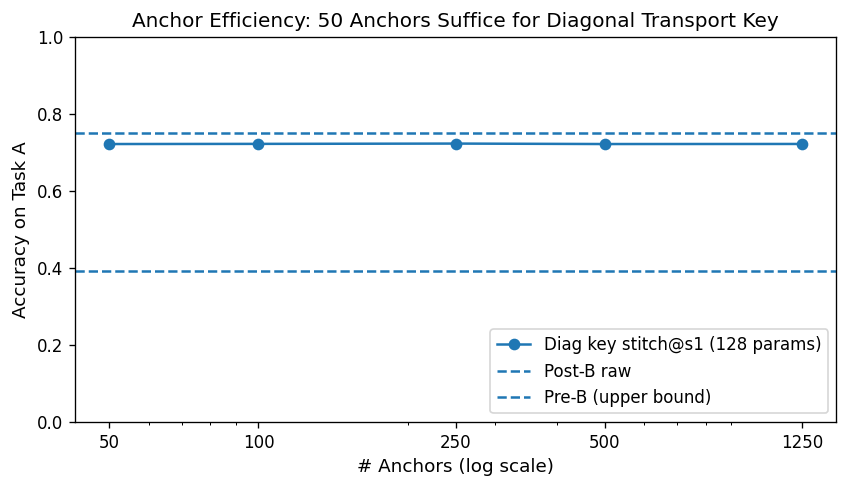}
    \caption{Anchor-efficiency analysis for Split CIFAR-100. Recovery reaches its effective plateau with a small balanced calibration set, suggesting that the relevant drift is low-dimensional or otherwise structured. The plateau indicates that the key is identifying a structured interface mismatch rather than relearning the task.}
    \label{fig:anchor_sweep}
\end{figure}

Figure~\ref{fig:anchor_sweep} reports the sensitivity of Task A recovery to the amount of Task A calibration data used to estimate the transport key. The important qualitative result is that recovery saturates rapidly: after a small balanced set of Task A examples, adding substantially more examples produces little additional improvement. This supports the claim that the dominant failure mode is a structured interface shift rather than a high-capacity relearning problem.

The graph should be read as follows. The horizontal axis varies the amount of calibration evidence available for key estimation. The vertical axis reports recovered Task A accuracy under the stitched evaluation. A steep early rise followed by a plateau means that the post-update network already retains much of the Task A computation; the key only needs enough evidence to identify the interface mismatch. If the method were simply retraining Task A, we would expect a stronger dependence on additional examples.

\section{Control Experiments Under Domain Shift}

\begin{figure}[h]
    \centering
    \includegraphics[width=0.85\linewidth]{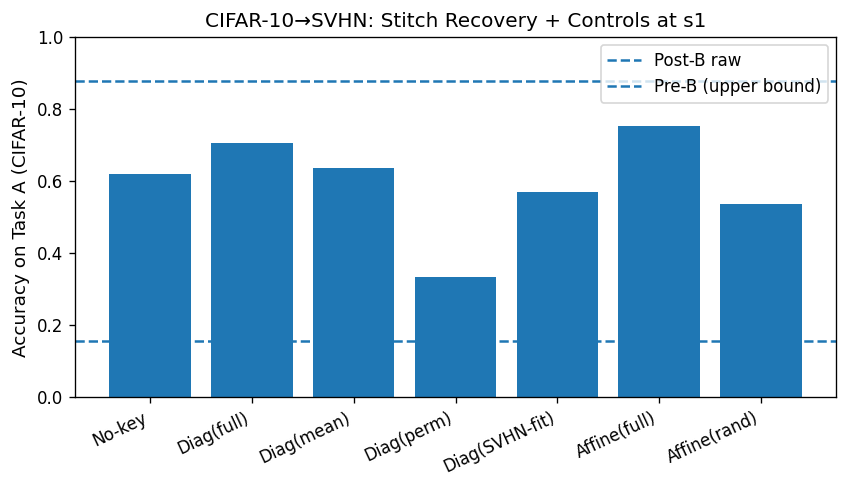}
    \caption{Control analysis at stage 1 for CIFAR-10$\to$SVHN. The transport key substantially improves stitched Task A accuracy, while controls that disrupt the relevant alignment structure reduce the gain.}
    \label{fig:controls_s1}
\end{figure}

\begin{figure}[h]
    \centering
    \includegraphics[width=0.85\linewidth]{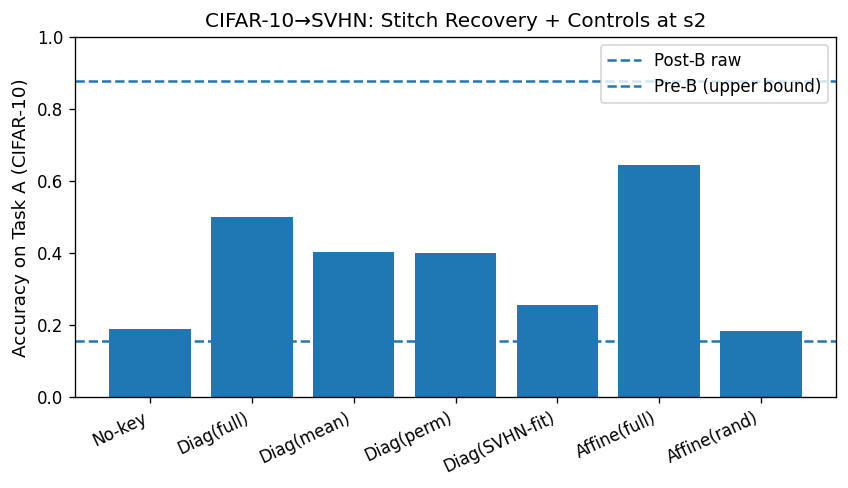}
    \caption{Control analysis at stage 2 for CIFAR-10$\to$SVHN. The valid keyed condition remains substantially above the post-update baseline, while correspondence-breaking controls collapse toward the forgotten model.}
    \label{fig:controls_s2}
\end{figure}

Figures~\ref{fig:controls_s1} and~\ref{fig:controls_s2} expand the CIFAR-10$\to$SVHN control analysis from Table~\ref{tab:domain_controls}. This setting is intentionally more difficult than the same-domain CIFAR split: after learning SVHN, the model's preserved CIFAR-10 head receives internal representations that have shifted under a different visual distribution. The controls ask whether recovery comes from meaningful interface alignment or from a generic improvement produced by inserting an additional module.

The stage-1 graph in Figure~\ref{fig:controls_s1} compares keyed recovery against structure-disrupting controls. The transport key recovers a large fraction of the lost CIFAR-10 accuracy, while disrupting the relationship between the key and the intended interface sharply reduces performance. This pattern is important: it shows that the improvement depends on preserving the correspondence between the post-update representation and the pre-update decoder.

The stage-2 graph in Figure~\ref{fig:controls_s2} shows a sharper version of the same phenomenon. At this later interface, simple compatibility is weaker, so the difference between valid keyed alignment and disrupted controls becomes more diagnostic. The result supports the interpretation that recovery is not an artifact of the classifier head or a generic smoothing effect. It depends on the internal geometry of the source and target interfaces.

\section{How to Interpret the Stage-Wise Results}

Across the main and supplementary figures, the same stage-wise pattern recurs: early interfaces exhibit the strongest recovery, middle interfaces show partial recovery, and late interfaces often fail. This is consistent with the access-based view of forgetting. Early and mid-level computation remains partially reusable after sequential training, but the coordinate system or interface expected by the original downstream model has shifted. Later representations are more task-specific and more strongly rewritten by subsequent training, leaving less recoverable structure for stitching.

This interpretation also explains why no-key stitching can sometimes perform well, especially in the Mini ViT experiment and the first stage of the domain-shift experiment. In those cases, the post-update early representation remains partly compatible with the pre-update decoder even without a key. The transport key is most informative when no-key stitching fails but keyed stitching succeeds, because that pattern isolates interface drift from direct feature preservation.

\section{Methodological Takeaway}

Taken together, the supplementary graphs support the central mechanism proposed in the paper. Small anchor sets are sufficient because the model is not relearning Task A from scratch; controls degrade performance because the key depends on real correspondence between old and new internal representations; and stage-wise degradation shows that recoverable latent structure is concentrated earlier in the network. This combination of anchor efficiency, control sensitivity, and stage localization is the empirical signature of transport-key recovery.

\end{document}